\documentclass[journal, twocolumn]{IEEEtran}
\IEEEoverridecommandlockouts

\usepackage{cite}
\usepackage{titlesec}
\usepackage{float}
\usepackage{graphicx}
\usepackage[acronym]{glossaries}
\usepackage{booktabs}
\usepackage[colorlinks=true, allcolors=blue]{hyperref}

\ifCLASSINFOpdf
\else
\fi
\begin{document}
%

\title{\large Sensor Integration and Performance Optimizations for Mineral Exploration\\ using Large-scale Hybrid Multirotor UAVs 
}


\author{
Robel Efrem$^{\dagger}$, 
Alex Coutu$^{\star}$,  
Sajad Saeedi$^{\dagger}$
\vspace{-7 mm}
\thanks{\noindent$^{\dagger}$Toronto Metropolitan University\quad\quad $^{\star}$Rosor Exploration \newline \textcolor{white}{---} \{robel.efrem, s.saeedi\}@torontomu.ca, alex@rosor.ca}%
}

\maketitle

\begin{abstract}
In this paper, the focus is on improving the efficiency and precision of mineral data collection using UAVs by addressing key challenges associated with sensor integration. These challenges include mitigating electromagnetic interference, reducing vibration noise, and ensuring consistent sensor performance during flight. The paper demonstrates how innovative approaches to these issues can significantly transform UAV-assisted mineral data collection. Through meticulous design, testing, and evaluation, the study presents experimental evidence of the efficacy of these methods in collecting mineral data via UAVs. The advancements achieved in this research enable the UAV platform to remain airborne up to 6$\times$ longer than standard battery-powered multirotors, while still gathering high-quality mineral data. This leads to increased operational efficiency and reduced costs in UAV-based mineral data-gathering processes.

\end{abstract}


%
\IEEEpeerreviewmaketitle


\section{Introduction}

The global economy heavily relies on mineral exploration, a process traditionally hindered by slow, labor-intensive ground-based methods subject to environmental constraints. The advent of UAVs (Uncrewed Aerial Vehicles) introduces a new paradigm in data collection techniques, promising to revolutionize this field by offering efficient and effective access to remote and previously unreachable terrains \cite{leech2021acquisition}.

Integrating high-sensitivity sensors onto UAV platforms presents significant challenges, including managing electromagnetic interference (EMI) and ensuring stable sensor performance across diverse environmental conditions. These technical obstacles are critical barriers to refining UAV-assisted mineral data collection processes \cite{mohsan2023editorial, vangu2022use}.

Current state-of-the-art approaches in UAV-based mineral exploration leverage gas hybrid multirotor UAVs for their high payload capacity and extended endurance, facilitating extensive coverage over large and remote areas. However, existing methodologies still grapple with issues like EMI management and sensor stabilization, indicating a gap in achieving optimal operational efficiency \cite{leech2021acquisition}.

This paper proposes a novel approach employing gas hybrid multirotor UAVs equipped with innovative strategies to mitigate the aforementioned challenges, thereby enhancing mineral data acquisition. The contributions include the development of advanced techniques for EMI management and sensor performance stabilization, significantly extending UAV airborne time and improving data quality over extensive terrains. 
The platforms acquired data will be compared directly to a dataset gather by Piper Navajo fixed wing aircraft typically used commercially in the mineral acquisition space. Fig.~\ref{fig:Suspended VLF} shows the platform developed to carry a suspended Very Low Frequency (VLF) electromagnetic (EM) and magnetometer sensor system. See videos of the flight on the website of the project\footnote{\href{https://sites.google.com/view/mineral-aircraft/multirotor-uav}{https://sites.google.com/view/mineral-aircraft/multirotor-uav}}.

The rest of the paper is organized as follows. Sec.~\ref{sec:rev} presents the literature review. 
Sec.~\ref{sec:method} describes the method. 
Sec.~\ref{sec:exp} shows the experimental results.  
Sec.~\ref{sec:con} concludes the paper.

\section{Literature Review} \label{sec:rev}
The integration of UAVs into the domain of mineral exploration represents a significant technological shift, enhancing the efficiency and scope of geophysical surveys. This literature review delves into the evolution of survey methodologies, focusing on the deployment of hybrid UAV systems equipped with advanced sensor technologies for mineral sensing. 

\begin{figure}[t!]
    \centering
    \includegraphics[width=7.4cm]{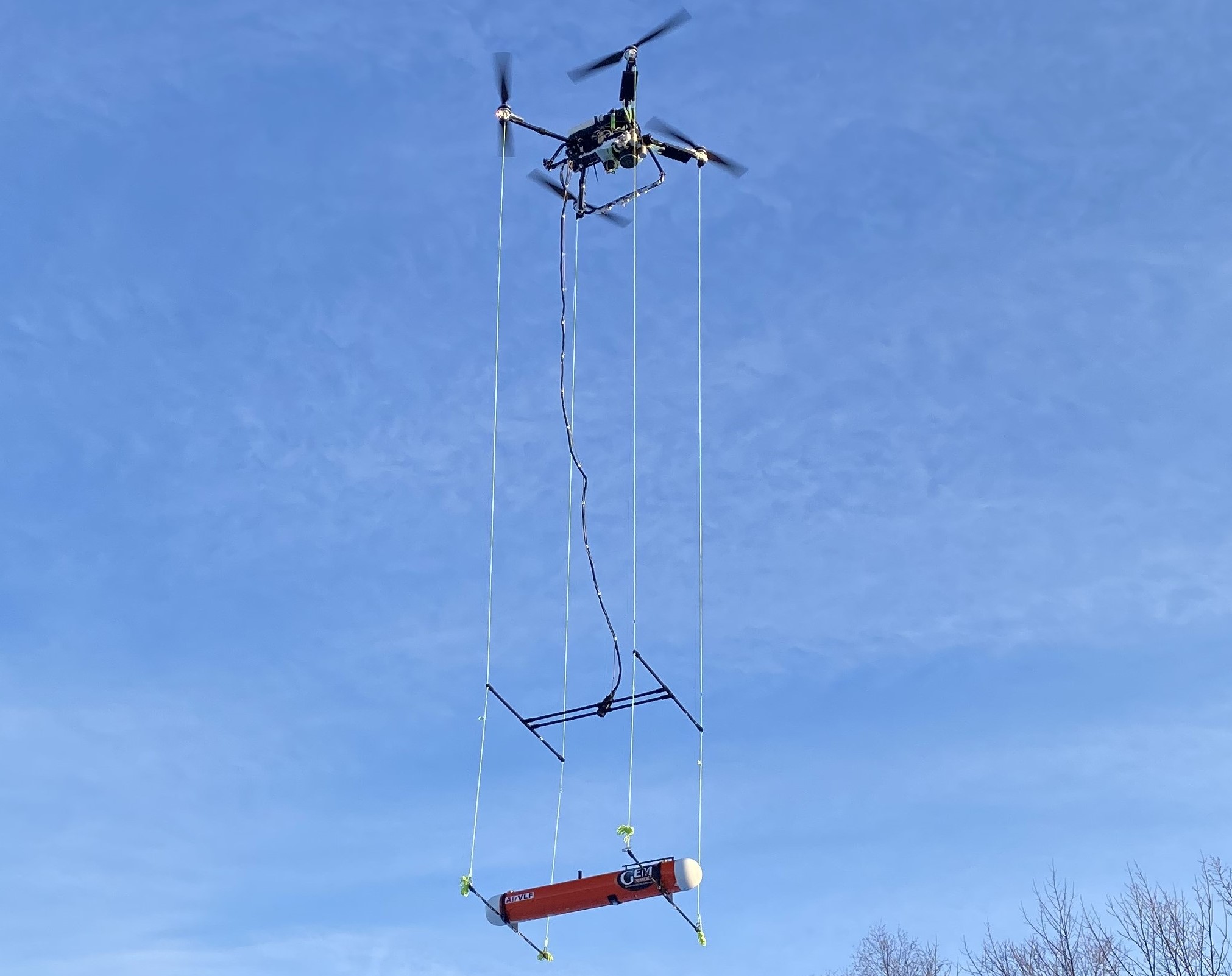}
    \caption[Suspended multiple payload system for VLF EM]{Suspended multiple payload system, VLF EM (bottom) and magnetometer (above) utilizing a four-cable suspension system while in flight.}
    \label{fig:Suspended VLF}
\end{figure}

Historically, the identification of subsurface mineral deposits has relied heavily on traditional ground-based magnetic surveys. These surveys utilize magnetometers to detect anomalies in the Earth's magnetic field, indicative of mineral deposits \cite{vitale2019new}. Despite their effectiveness, such methods are labour-intensive and often fall short of covering vast or inaccessible terrains. The introduction of aerial surveys, particularly those conducted via helicopters, marked a significant improvement by providing higher data resolution in smaller or more complex areas. However, these advancements came with increased operational costs \cite{persova2021resolution}.

The advent of UAV technology in mineral exploration has heralded a new era of survey methods. UAVs, with their ability to operate at lower altitudes, offer substantial benefits for conducting detailed surveys over small or complex terrains. These benefits include enhanced safety through the minimization of manned flights and the provision of high-resolution data, although they are somewhat constrained by factors such as flight duration, control range, and payload capacity \cite{jiang2020integration, greengard2019when}. The emergence of long-endurance hybrid UAVs, designed for low-altitude, high-endurance flights, presents a promising solution for efficient and cost-effective mineral data acquisition over extensive areas \cite{yang2022hybrid, saeed2018survey}. An example Hybrid UAV with VLF sensor is shown in Fig.~\ref{fig:Suspended VLF}.

Further advancements in sensor technology have expanded the utility of UAVs in mineral detection. The incorporation of hyperspectral imaging and LiDAR sensors into UAV systems has enabled detailed surface analysis and precise topographical mapping, essential for distinguishing between rock types and identifying geological formations \cite{barton2021extending, niethammer2012uav, salvini2015geological}. Moreover, the combination of different sensor types, such as magnetic sensors with hyperspectral imaging or LiDAR with multispectral imaging, has significantly enhanced the capability of UAVs in mineral exploration \cite{jackisch2019drone, shendryk2020fine}. The integration of heavier sensors like VLF or radiometric sensors on long-range platforms is an area ripe for exploration, aiming to achieve high efficiency in mineral data acquisition.

Despite these technological strides, challenges such as signal-to-noise ratio, data resolution, sensor calibration, and data fusion remain. These challenges are particularly pronounced with advanced sensors like VLF, magnetometers, hyperspectral, and LiDAR, underscoring the need for further research in these areas \cite{goetz2009three, vosselman2013recognising}. Additionally, the complexity of integrating and interpreting data from multiple sensors on a single UAV platform presents substantial challenges \cite{thiele2021multi}. The regulatory landscape governing UAV usage in mineral exploration also varies significantly, affecting their deployment and operational capabilities \cite{molnar2016unmanned}. 

The integration of VLF sensors on UAVs, demonstrated in this paper as well as through UAV-electromagnetic systems development, showcases the advancement in aerogeophysical methods \cite{eross2013three}. These systems enable extensive geophysical mapping and detailed electromagnetic soundings over large areas and complex terrains, detecting subsurface features like pipelines and achieving depth soundings beyond 300 meters \cite{parshin2021lightweight}. 
The adaptability of these UAV systems for simultaneous magnetic and VLF surveys underscores their versatility in addressing a wide spectrum of geological challenges, enhancing the scope of UAV applications in geophysical exploration. This signifies a leap in mineral exploration efficiency, blending lightweight design with high-frequency operational capabilities for comprehensive geological assessments.

\textcolor{black}{Our approach leverages the advancements in hybrid UAV and sensor technologies to address the existing challenges in mineral exploration. By integrating long-endurance hybrid UAVs with advanced sensors like VLF, magnetometers, and radiometric sensors, we aim to improve the signal-to-noise ratio, enhance data resolution, and streamline UAV implementation processes. Our methodology is specifically designed to overcome the limitations of payload capacity and flight duration, enabling comprehensive and efficient mineral data acquisition over vast and inaccessible terrains using hybrid UAVs. Furthermore, our system incorporates a dual payload system for the potential analysis of multi-sensor data, thereby addressing the complexities associated with the current UAV-based exploration methods. This holistic approach not only mitigates the operational and regulatory challenges but also sets a new standard for cost-effectiveness and environmental sustainability in the mineral exploration industry.}

\section{Proposed Methods}\label{sec:method}
In this section, we will present the proposed strategies aimed at tackling issues such as electromagnetic interference and payload dynamics. 

\subsection{VLF EM Integration to Gas Hybrid Platform}
The advent of gas hybrid UAV platforms, known for their enhanced payload capacity and endurance, has made it feasible to conduct long-range mineral data acquisition with heavier payloads. The recent progress in the miniaturization of VLF EM sensors has facilitated their integration into UAV systems.

\subsubsection{VLF EM Functionality}
\textcolor{black}{A VLF EM sensor is a type of geophysical instrument used on drones to detect variations in electromagnetic fields beneath the surface of the Earth, mainly for mineral exploration and the detection of underground structures. It operates by receiving naturally occurring or artificially generated VLF electromagnetic waves, which penetrate the ground and interact with subterranean conductive materials. These interactions create secondary electromagnetic fields that the sensor detects, allowing for the mapping of geological features or mineral deposits. Integrating VLF EM sensors onto drones presents several challenges, including the sensor's sensitivity to the drone's orientation, altitude changes, and EMI, which can affect measurement accuracy. Additionally, the size and weight of VLF EM sensors can limit the operational endurance and maneuverability of the drone as they are typically in excess of 6kg, the electromagnetic interference from the drone's own electronic systems can further complicate data acquisition and interpretation.}

\subsubsection{VLF EM Noise Threshold Identification}

The first step in this integration involves assessing the noise level generated by the UAV to establish an optimal distance from the drone, ensuring that the UAV's operational noise does not compromise data integrity. EMI and Radio Frequency Interference~(RFI) represent disruptions caused by external electromagnetic sources in electrical circuits. In UAVs, sources of EMI include electric motors, power supplies, electronic speed controllers, and other onboard electronics, with the primary source being the gasoline AC generator. These components can generate electromagnetic fields during UAV operation, potentially interfering with the VLF EM sensor and leading to false readings, thereby compromising data accuracy and reliability.

To mitigate EMI in a VLF EM system mounted on UAVs, strategies similar to those used for magnetometer mounting on UAVs are employed. This involves physically distancing the VLF EM sensor from the UAV. Options include mounting the sensor on an extended boom or suspending the payload to increase the separation between the sensor and the EMI source. However, the considerable weight and size of typical VLF EM sensors render the use of booms or masts impractical, as they negatively impact the UAV's flight performance and stability.

To determine the noise threshold at which VLF EM data becomes unreliable, several tests were conducted. The initial tests involved setting up the VLF EM sensor with all ancillary equipment as though it were mounted on a drone. This assembly was then placed on a cage-like structure 2 meters above the ground. The UAV, flying at approximately 3 m/s, passed over the stationary sensor at various angles and altitudes to assess noise interference in the data. Another series of tests involved the UAV hovering and yawing in place over the sensor, executing four complete rotations at different heights. Through these tests, an altitude of 9 meters was determined to be the threshold at which acceptable noise levels were achieved, with altitudes ranging from 4 to 15 meters being explored.

\subsubsection{VLF EM Mounting}

With the optimal separation distance determined, a crucial aspect of the integration process was the development of a unique four-point mounting system with stabilizers for the suspended payload configuration, as illustrated in Fig.~\ref{fig:Suspended VLF}. While traditional suspended magnetometry employs a single-point mounting system for simplicity, VLF EM sensors are highly sensitive to changes in heading. \textcolor{black}{These sensors are typically sensitive to heading changes during flight because their measurements are influenced by the orientation of the sensor relative to the Earth's magnetic field, leading to variations in the detected electromagnetic signals based on the sensor's alignment \cite{eross2013three}.} Without a method to maintain the heading in line with the UAV, data quality would suffer. While some sensors come with a built-in empennage to align with the flight direction, these systems have drawbacks such as settling time, susceptibility to crosswinds, and poor performance at low speeds. To address these challenges, a suspension system was designed to attach to the four motor arms of a quadrotor and extend downward to a platform equidistant from the sensor. This configuration ensures that the payload remains aligned with the UAV's heading, unaffected by wind conditions or flight speed variations. The system utilizes quick connects to secure a 9-meter length of paracord below each of the four motors. These four cable lengths are connected to a payload mount extending outward at equal distances from the motor arms, effectively creating a 4-bar linkage that keeps the payload level during flight.

Furthermore, an intermediate lightweight platform was developed to enhance yaw stability. As the cable length increases, yaw performance deteriorates. The intermediate platform adds rigidity to the suspension system and prevents cable entanglement during UAV yawing, minimizing payload lag time in response to yaw input changes. It was designed to minimize payload swing and act as a secondary payload mount, minimizing relative movement between the two suspended sensors. These factors are essential for preserving the integrity of the collected data. This secondary level not only enhances yaw stability but also reduces payload swing and settling time, allowing for shorter lead-in and lead-out distances for each survey line. Consequently, this reduces the overall time the UAV spends in the air for the same amount of production, thereby increasing the system's productivity without altering the UAV's flight characteristics.


\subsection{Vibration Analysis}
The sensors, tasked with the detection and quantification of natural gamma radiation from geological substances, are inherently sensitive and demand a stable environment to ensure precise data capture. The integration of these sensors on durable, gasoline-powered UAVs often introduces significant vibrational disturbances from the engine and propellers, corrupting the collected data. To counteract this, a specialized vibration damping solution, optimized for gasoline-powered UAVs used in radiometric surveys, was engineered. Its core aim was to create a secure and stable platform for the sensor, effectively shielding it from the engine and rotor-induced vibrations.

\begin{figure}[t!]
    \centering
    \includegraphics[width=8.5cm]{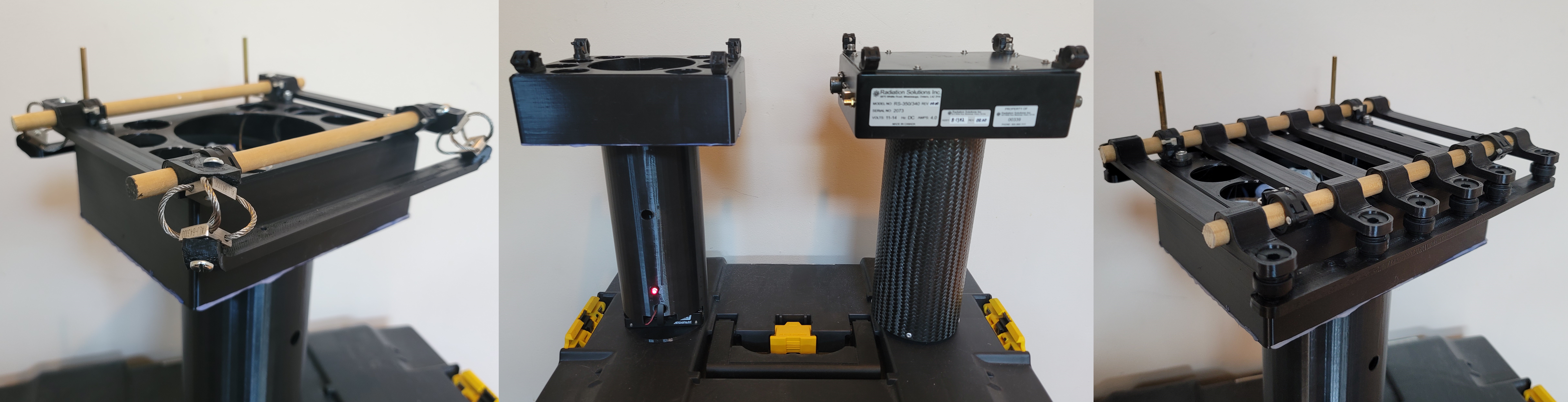}
    \caption[Two styles of vibration isolators with dummy payload]{Two styles of vibration isolators retrofitted to 3D printed mock vibration measurement payload (left wire rope isolators and right rubber ball dampers), also real radiometric sensor (center right) vs mock payload (center left) comparison.}
    \label{fig:Vibration mount}
\end{figure}

Two prevalent isolator varieties employed in this context are metal wire rope isolators and rubber ball dampers, illustrated in Fig.~\ref{fig:Vibration mount}. Both types exhibit unique properties and efficacy in vibration attenuation. Metal wire rope isolators comprise intricately wound stainless steel wires, utilizing their elastic properties for flexibility and strength. This design enables them to absorb and dissipate vibrational energy omnidirectionally. Their key benefit lies in their resilience, capable of withstanding extreme environmental conditions, including temperature variations and chemical exposure, with minimal performance impact. Conversely, rubber ball dampers leverage the elastic deformation of their material, which reverts to its original shape post-deformation. This characteristic facilitates the absorption and dispersion of vibrational energy. Rubber ball dampers excel in their customization potential and ease of integration in various designs, although they are less resistant to environmental factors. Both isolator types, despite their differing mechanisms, are fundamentally geared towards mitigating vibrations across a broad frequency spectrum.

The development of this vibration damping solution included a series of theoretical evaluations and simulations aimed at determining the most effective combination of damping materials, mounting strategies, and structural modifications to minimize vibration transfer to the radiometric sensor. These theoretical plans were subsequently put to the test through empirical trials with various configurations, assessing their vibration reduction capabilities.

Determining the most suitable damper set for this specific application necessitated the use of a damping effectiveness equation. In Eq.~\eqref{ed:damp}, $D$ symbolizes the damping effectiveness, quantified as the percentage decrease in the vibration's amplitude. The parameter $\eta$ stands for the initial vibrational intensity, expressed in acceleration units (m/s$^2$), $\zeta$ denotes the damping ratio of the isolators, a dimensionless quantity, and $S$ signifies the isolators' stiffness, measured in N/m. The payload mass connected to the damping system, denoted by $m$, is measured in kilograms, $n$ indicates the number of isolators in the system, and $f$ represents the vibrational frequency, in hertz (Hz) ~\cite{Halliday2021}:

\begin{equation}
    D = \frac{\eta \zeta S}{m  n  f^2}.
    \label{ed:damp}
\end{equation}

Upon completion of the calculations, it was determined that the ideal damping arrangement would consist of four metal wire rope isolators. These isolators, positioned at 45 degrees at each corner of the payload, were identified as providing the optimal damping effect for the system's tuning. In contrast, the rubber ball damper mounts were designed to accommodate a range of 4 to 12 isolators, with 6 being the theoretically optimal number. This variability allowed for empirical testing beyond the theoretically ideal configurations, ensuring the validity of the chosen setup.

The testing methodology involved a comprehensive 3D modeling of a sensor replica, inclusive of a vibration logging apparatus. This replica, mirroring the Radiation Solutions RS-530 in weight distribution and standard mounts, was pivotal in accurately simulating the payload's in-flight oscillations. Embedded within this test model were a microcontroller for vibration logging, gyroscopes, and accelerometers for detailed post-flight analysis. The model also incorporated an integrated battery and a LED indicator for logging activity. The completed test payload, juxtaposed with the reference payload, is depicted in Fig.~\ref{fig:Vibration mount}.

\begin{figure}[t!]
    \centering
    \includegraphics[width=8.5cm]{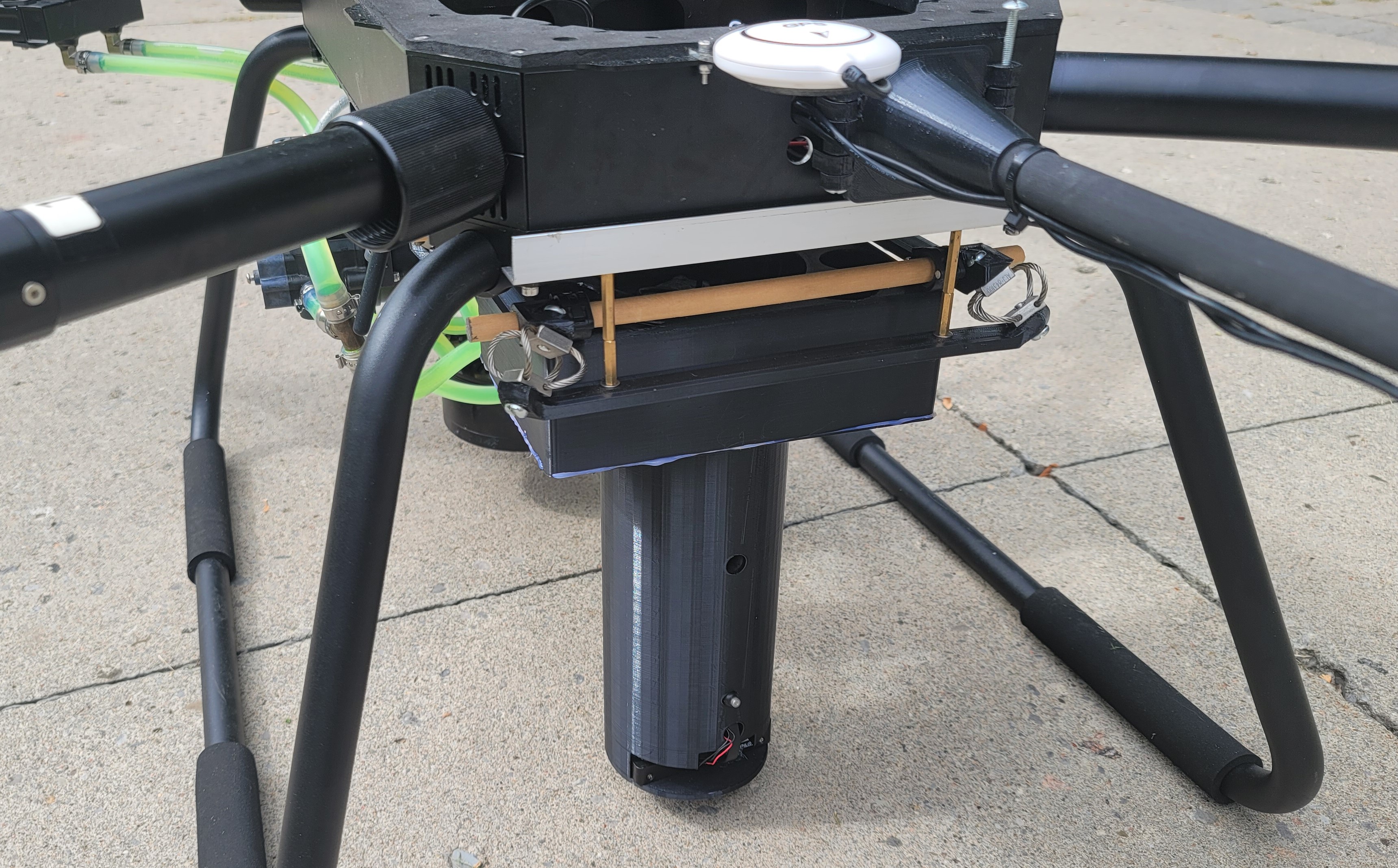}
    \caption{Example testing configuration for metal wire vibration isolators in optimal positioning and orientation.}
    \label{fig:Mount on UAV}
\end{figure}

Subsequently, the CAD representation allowed for the design of two distinct vibration isolation systems, facilitating swift evaluation of their vibration damping properties. Keeping in mind the specific payload characteristics, specially tailored metal wire rope isolators were procured, alongside a set of rubber damping balls suited for the payload's weight. The mounting design for both isolators aimed for compactness while ensuring that the isolators functioned under compression, as this method yields a more consistent deformation than tension, leading to predictable and reliable responses and avoiding material overstretching.
The experimental phase involved attaching the newly designed vibration isolation mounts to the base of the hybrid UAV as shown in Fig.~\ref{fig:Mount on UAV}. A series of flight tests, ranging from gentle hovering to dynamic maneuvers, were conducted to test the vibration damping efficiency of each configuration. These tests, lasting a minimum of one minute each, provided sufficient data for comprehensive post-flight analysis.

\begin{figure}[t!]
    \centering
    \includegraphics[width=8.5cm]{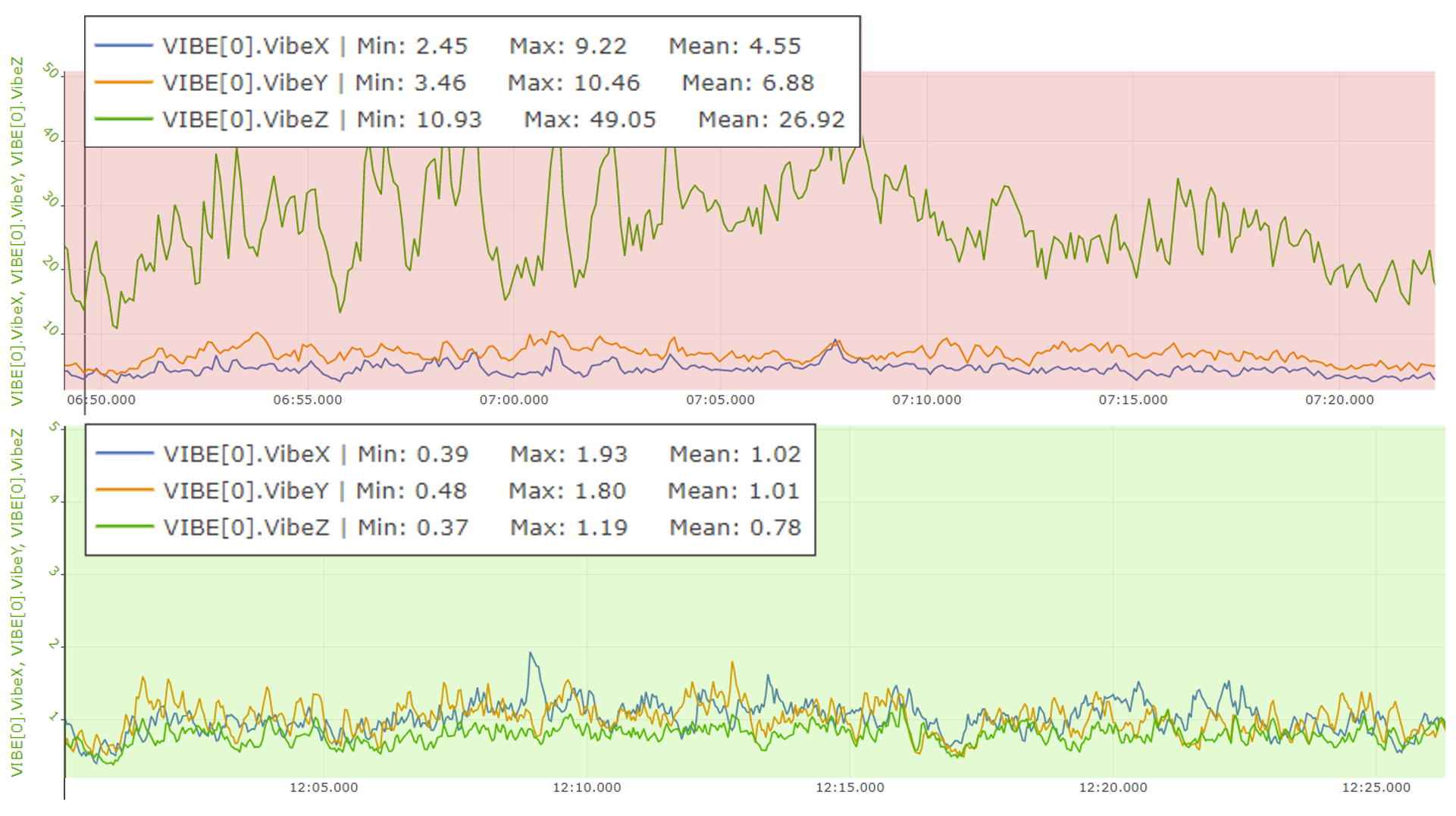}
    \caption{Initial vibrations present in frame (top) measured from payload, and Vibrations measured in payload after damping (bottom) in $m/s^{2}$.}
    \label{fig:Vibration results}
\end{figure}

The results of the tests demonstrated a marked improvement in the reduction of unwanted vibrations across all axes, particularly on the Z axis. The experimental data aligned closely with the theoretical predictions, especially for the rubber ball dampers, which showed an estimated reduction in vibration amplitude of 97\%. While metal wire rope isolators effectively diminished vibrations to an average of $1.71 \, m/s^{2}$, a 23-fold reduction, the simplest setup using just four rubber ball dampers surpassed this, achieving a 35-fold reduction in Z-axis vibrations. This made it the most suitable configuration for this payload. The comparison data in Fig.~\ref{fig:Vibration results} illustrate a stark reduction in the vibrations on the Z-axis and notable decreases on the X and Y axes. The vibration attenuation, when quantified in decibels (dB), indicates a reduction of approximately 30.88 dB for a 35-fold decrease. Analysis of the overall aircraft noise spectrum revealed dominant harmonics at approximately 33HZ and a secondary at around 76HZ, attributable to engine, propellers, motors, and frame resonance, as recorded from the aircraft's central point during flight.

In conclusion, the initial calculations and empirical findings led to the design and implementation of an effective vibration-damping platform. This platform successfully isolated the radiometric sensor from the vibrations produced by the gasoline engine, thereby enabling stable and accurate data collection. Field surveys using this system confirmed its suitability for radiometric data acquisition on a UAV platform. The vibration-damping solution's ability to minimize noise-induced errors significantly enhances the quality and accuracy of the collected radiometric data, paving the way for more precise and reliable mineral exploration using UAV technology in various challenging environments.

\section{Results} \label{sec:exp}
\textcolor{black}{This section presents the findings from our comprehensive surveys using a gas hybrid system equipped with a VLF EM sensor and radiometric sensor.}
\subsection{Gas Hybrid System VLF Survey Results}
The phenomenon known as an interference halo is observed when wave patterns overlap and interact, leading to a disturbance or alteration in the electromagnetic signals around the equipment \cite{walter2021optimization}. Investigations of the VLF EM capabilities of a hybrid UAV system revealed the extent of its electromagnetic interference halo at high frequencies. During preliminary static evaluations, the halo's dimensions were determined by evaluating the interference at each consecutive distance revealing a consistent 7m radius where the signal's amplitude was slightly above the ambient noise. It was found that elevating the sensor payload to 8m above the ground could significantly reduce the interference experienced during flight operations. Conversely, maintaining a 6m altitude would result in interference levels of 3-4\%, which is substantially higher than the optimal noise floor resolution, based on comparisons of signal interference amplitude against a background noise level of $\pm 0.2$ nT detected in preliminary buzz tests. \textcolor{black}{A buzz test is a practice where the UAV sensor is mounted on a platform and the UAV is flown above the sensor at a variety of altitudes to measure the noise generated at each altitude, this will show what suspension length allows for a reasonable amount of noise}. Extending the suspension length of the magnetometer to 9m was effective in diminishing the interference signals, although it's important to note that these noise levels do not alter the frequency of VLF readings but instead affect their amplitude by causing fluctuations around the baseline signal level.

The VLF EM surveys carried out in Hagersville, Ontario, Canada yielded data of exceptional quality and stability throughout the flights. Analysis of the UAV flight log demonstrated that pitch and roll variations were kept within a tight range of $\pm5^\circ$, without any detectable interference between the magnetometer and VLF EM sensors despite their proximity of just 1m. These surveys were methodically executed over the site in a grid pattern at varying heights (5m, 22m, and 52m above ground level), focusing particularly on the noise levels at the lowest elevation to assess the VLF EM data's consistency across different channels. The uniformity observed across all channels at every survey grid level highlights the system's stability, with special attention given to the survey conducted at 5 meters above ground level for detailed analysis.

\begin{figure}[t!]
    \centering
    \includegraphics[width=8.5cm]{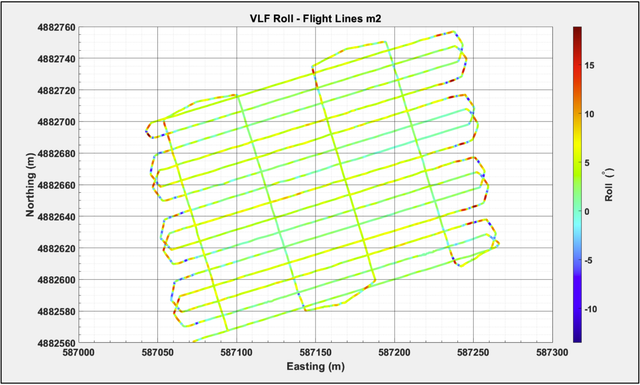}
    \vspace{-2mm}
    \caption[VLF-EM roll variations]{The VLF-EM roll variations for the unprocessed m2 flight data.}
    \label{fig:roll variations}
\vspace{2mm}
    \centering
    \includegraphics[width=8.5cm]{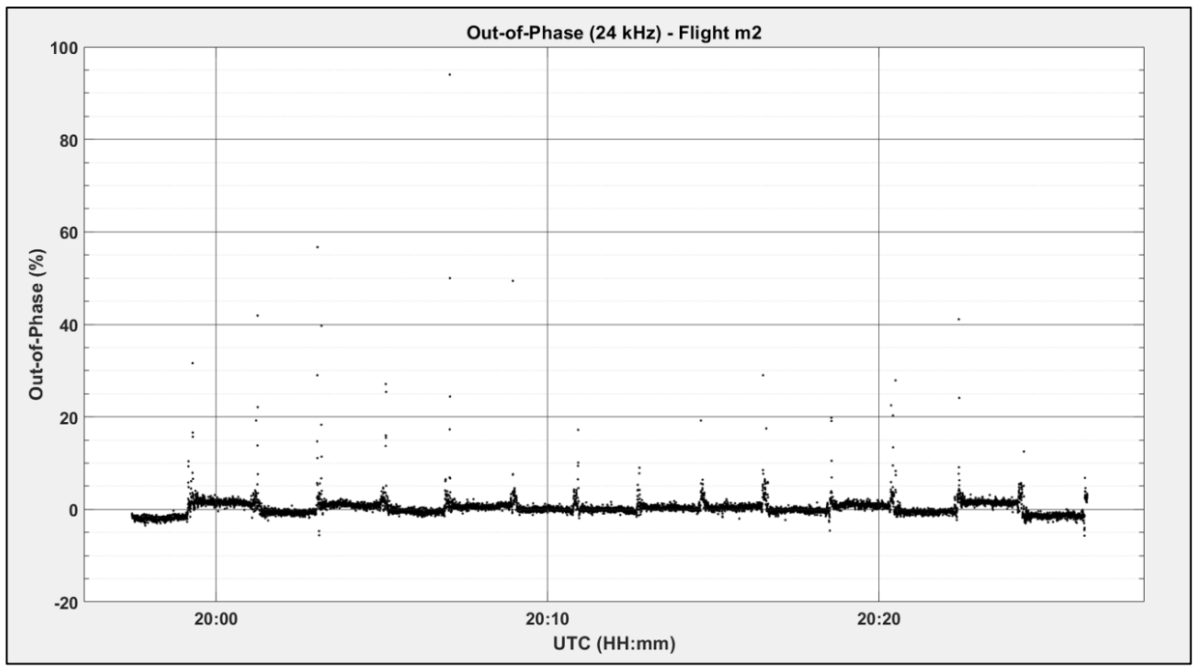}
    \vspace{-2mm}
    \caption[Out-of-phase (24 kHz) VLF EM variations]{Profile of the unprocessed out-of-phase (24 kHz) VLF EM variations}
    \label{fig:Out of phase}
\vspace{2mm}
    \centering
    \includegraphics[width=8.5cm]{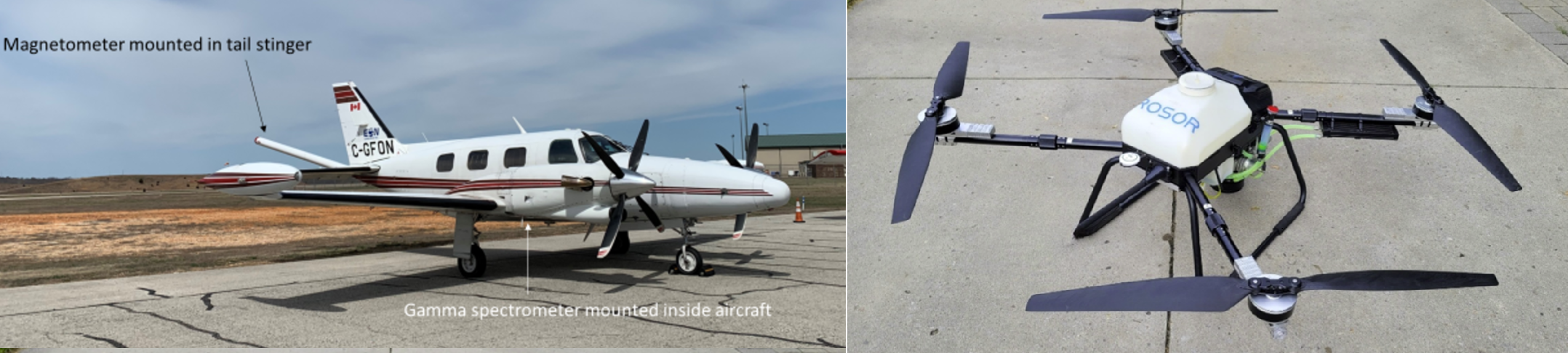}
    \vspace{-2mm}
    \caption[Hybrid UAV vs a Piper PA-31 Navajo]{A Piper PA-31 Navajo with integrated magnetometer and internal Gamma-ray spectrometer (right) compared to the hybrid UAV system used (left). How the radiometric sensor is mounted shown in Fig.~\ref{fig:Mount on UAV} \cite{usgs2023bipartisan}}
    \label{fig:PiperVsUAV}
\end{figure}

A significant noise source in the collected data was the pendulum-like motion of the VLF-EM sensor, especially noticeable at the turn points of the survey lines, affecting primarily the in-phase and out-of-phase channels along with h1, h2 (horizontal components of the electromagnetic field), and pT \textcolor{black}{(total magnetic field strength)} channels. \textcolor{black}{The in-phase channel refers to the component of the detected electromagnetic signal that is aligned with the primary field's oscillation, indicating conductive materials, while the out-of-phase channel refers to the component that is delayed relative to the primary field, often indicating the presence of magnetic materials or the degree of permeability in the subsurface.} This analysis focuses on the data from the second of three VLF flights conducted at the Hagersville site, as it represented the longest duration of operation at the intended survey altitude. The roll variations, depicted in Fig.~\ref{fig:roll variations}, highlight the stability of the VLF-EM system in-flight, with most pitch and roll variations maintained within $\pm 5^\circ$ across survey lines. The absence of underground targets in the surveyed area meant that the refined data primarily reflected the system's inherent noise level, exhibiting minimal spatial variations. Analysis of the data channels, particularly the out-of-phase 25.2 kHz channel which was the weaker signal, indicated a final system noise level of 4\%, with the out-of-phase data channel's noise levels throughout the flight.

To enhance the interpretation of the VLF data, visual representation from the 24 kHz channel \textcolor{black}{(the channel with the highest penetration depth and resolution)} is provided through a side-on profile view, which more effectively showcases variations in noise levels compared to a grid-based approach. The out-of-phase channels illustrated in Fig.~\ref{fig:Out of phase}, exhibit notable variations primarily as a result of the swinging motion observed at the turns of the survey lines. However, within the straight segments of the survey, the out-of-phase channel consistently remains within acceptable noise boundaries, whereas the in-phase channel accurately mirrors the system's overall noise profile, exhibiting limited spatial discrepancies. The lack of significant anomalies within the collected data is due to the absence of intentionally introduced targets in the survey area. Overall, the noise levels across all data channels are considered to be within acceptable limits, indicating a satisfactory outcome for the survey's integrity.

\begin{figure}[t!]
    \centering
    \includegraphics[width=7cm]{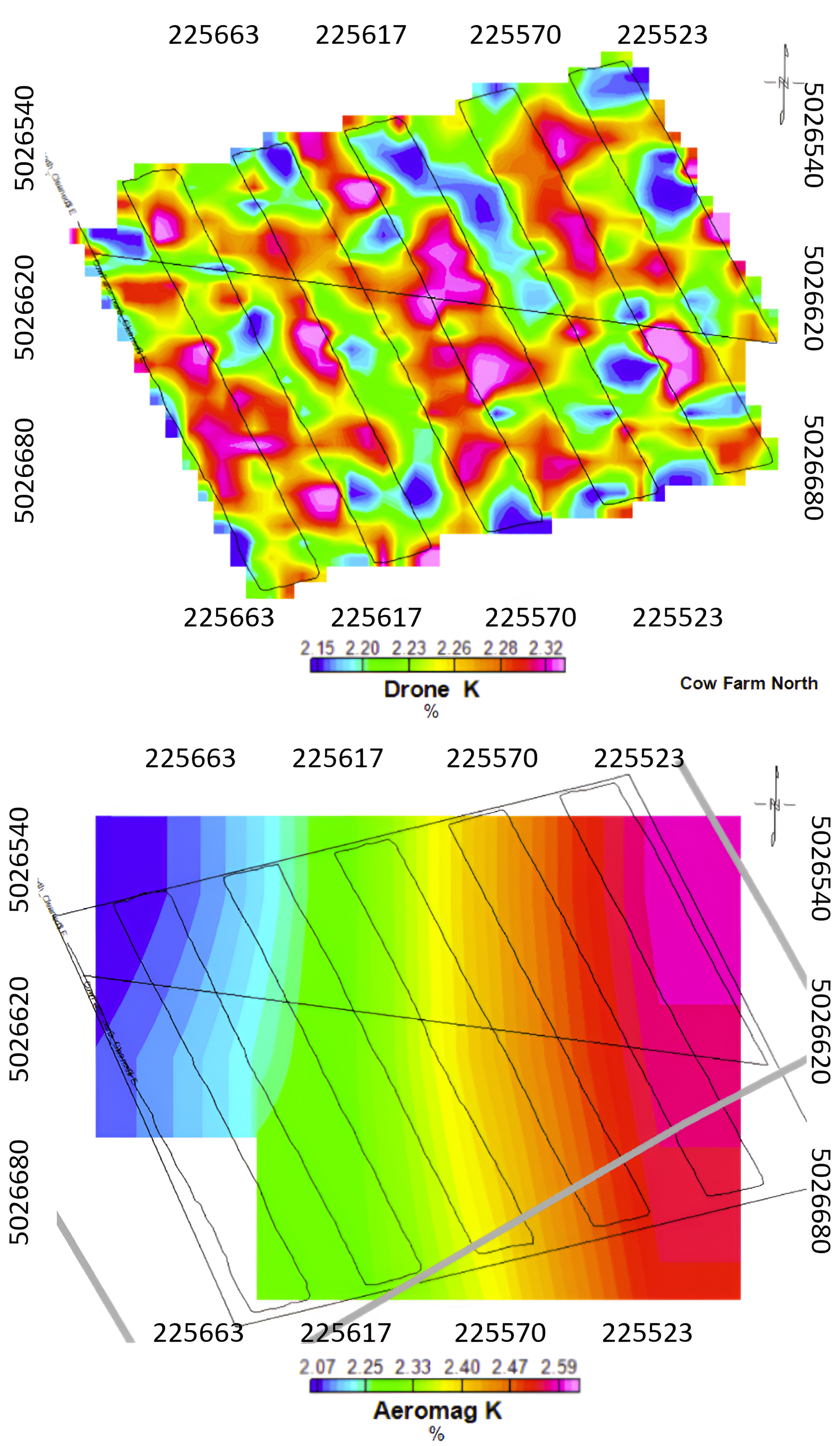}
    \caption[Experimental results of the radiometric survey]{Experimental results of the radiometric survey Potassium Comparison, drone survey (top) and fixed-wing survey (bottom) displaying the variations on a local scale of the potassium (K) distribution. Numbers surrounding the plot indicate Eastings and Northings. Both images contain overlays of the flight paths of both systems, the fixed-wing being the thicker gray lines and the drone survey being the thinner black lines.}
    \label{fig:Rad vs MAG.png}
\end{figure}
\begin{figure}[t!]
    \centering
    \includegraphics[width=7cm]{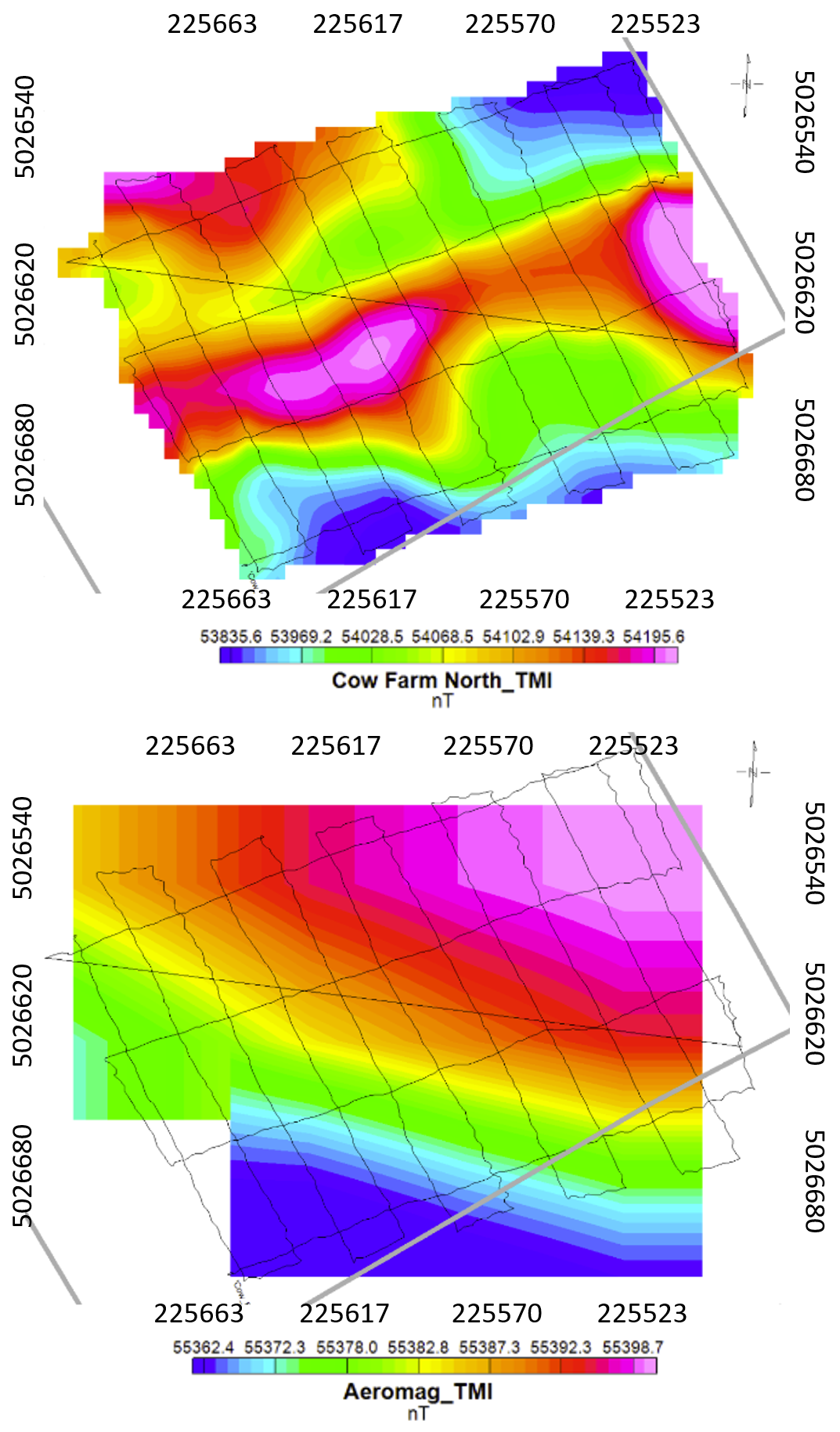}
    \caption[Experimental results of the magnetic survey comparison]{Experimental results of the magnetic survey comparison, drone survey (top) and fixed-wing survey (bottom) displaying the variations on a local scale of the TMI distribution. Numbers surrounding the plot indicate Eastings and Northings. Both images contain overlays of the flight paths of both systems, the fixed-wing being the thicker gray lines and the drone survey being the thinner black lines. The much larger fixed-wing dataset has been cropped to the drone survey size and as a result, the entire dataset is derived from two perpendicular flight lines. This results in much lower-resolution data.}
    \label{fig:Gasdronemag.png}
\end{figure}

\begin{table*}[t!]
\centering
\caption[Sample repeatability analysis of radiometric dataset]{\textcolor{black}{Sample repeatability analysis of radiometric dataset.} \textcolor{black}{The columns show the small differences between the sensor channels which are considered representative of
repetitive data}}
\resizebox{\textwidth}{!}{%
\begin{tabular}{cccccccccc}
\toprule
X (UTM) & Y (UTM) & Flights\_K (\%) & Tie\_K (\%) & Flights\_U (ppm) & Tie\_U (ppm) & Difference\_K (\%) & Difference\_U (ppm) \\
\midrule
327193 & 5030645 & 2.50 & 2.58 & 2.94 & 3.07 & -0.08 & -0.13 \\
327194 & 5030646 & 2.51 & 2.58 & 2.92 & 2.99 & -0.07 & -0.06 \\
327195 & 5030646 & 2.51 & 2.58 & 2.90 & 2.88 & -0.07 &  0.02 \\
327196 & 5030647 & 2.50 & 2.57 & 2.88 & 2.78 & -0.08 &  0.10 \\
327197 & 5030648 & 2.48 & 2.57 & 2.86 & 2.66 & -0.09 &  0.19 \\
327198 & 5030648 & 2.45 & 2.57 & 2.83 & 2.53 & -0.12 &  0.29 \\
327204 & 5030652 & 2.41 & 2.49 & 2.52 & 2.86 & -0.07 & -0.34 \\
327204 & 5030652 & 2.43 & 2.50 & 2.53 & 2.88 & -0.07 & -0.34 \\
327205 & 5030653 & 2.45 & 2.51 & 2.55 & 2.89 & -0.06 & -0.34 \\
327206 & 5030653 & 2.46 & 2.52 & 2.56 & 2.89 & -0.05 & -0.33 \\
327207 & 5030654 & 2.48 & 2.53 & 2.58 & 2.90 & -0.05 & -0.32 \\
327208 & 5030655 & 2.50 & 2.54 & 2.60 & 2.91 & -0.05 & -0.31 \\
327209 & 5030655 & 2.52 & 2.56 & 2.62 & 2.92 & -0.04 & -0.31 \\
327209 & 5030656 & 2.53 & 2.58 & 2.63 & 2.87 & -0.05 & -0.23 \\
327210 & 5030656 & 2.49 & 2.60 & 2.65 & 2.71 & -0.11 & -0.06 \\
327211 & 5030657 & 2.47 & 2.62 & 2.67 & 2.60 & -0.15 &  0.08 \\
\bottomrule
\end{tabular}
}
\label{table:repeat points}
\end{table*}

Given the satisfactory noise levels of the collected data, any anomalies identified in the dataset following a real-world survey can be confidently ascribed to subsurface geological formations, rather than being dismissed as artifacts of electromagnetic interference (EMI) or system-induced roll variations. These detected anomalies can subsequently be analyzed and visualized to assess their potential as mineral deposits, including the determination of the type of minerals present. This step is crucial for distinguishing genuine geological features from noise, thereby enabling accurate identification and characterization of mineral resources.

\subsection{Gas Hybrid Radiometric Survey Results}
Following the successful mitigation of vibrations from the onboard hybrid generator to levels acceptable for the specified radiometric sensor, the gas hybrid multirotor system was deemed ready for operational validation. This step was crucial to demonstrate the system's capability to carry out effective mineral sensing operations. The system's robust payload capacity enabled the concurrent execution of both radiometric and magnetometric surveys, showcasing its operational efficiency. This integrated approach allows for a comprehensive analysis, where the results derived from the UAV are juxtaposed with data from traditional manned aircraft surveys for a thorough comparative study. Specifically, the performance of this UAV system will be contrasted with data from a 2014 survey conducted using a Piper Navajo aircraft shown in Fig.~\ref{fig:PiperVsUAV}, focusing on discrepancies in operational specifications and capabilities.

The UAV system utilized in these tests was outfitted with advanced sensing equipment, including the Geometrics MagArrow magnetometer and the Radiation Solutions RS-530 gamma-ray spectrometer, to capture high-quality magnetic and radiometric data. 

Field tests were undertaken over agricultural land in the Renfrew Ontario area, with the focus on the largest field to provide a comprehensive evaluation of the UAV system's performance relative to manned aircraft surveys. The UAV followed a predefined grid pattern at a consistent altitude and speed, facilitating the simultaneous collection of magnetic and radiometric data. Following data acquisition at a remote agricultural site, a detailed quality control analysis of the radiometric data was initiated. Without standard radiometric Test Lines (predefined paths flown to verify and calibrate
the consistency and accuracy of the sensor) or Th source tests available, the UAV conducted additional survey lines to ensure the reliability and accuracy of the data processed using Noise-Adjusted Singular Value Decomposition (NASVD). \textcolor{black}{A Th source test refers to a test using a Thallium-204 (Tl-204) radioactive source to calibrate and test the sensor to ensure it is functioning correctly.}

This statistical technique is instrumental in improving the signal-to-noise ratio of the collected data, enhancing the overall quality of the survey results. The inclusion of tie lines, not typically necessary for radiometric surveys, served as a means to further validate the system's data accuracy and repeatability under consistent weather conditions. \textcolor{black}{A tie line is a line of data acquisition that crosses the primary survey lines perpendicularly, used to ensure data consistency and accuracy across the survey area. Analysis of these lines for potassium (K) and uranium (U) showed only subtle differences. Specifically, the variance in measurements between flight and tie lines for potassium ranged up to 0.15\%, while for uranium, it ranged up to 0.34 ppm, as detailed in Table~\ref{table:repeat points}. These results highlight the system's consistency in data acquisition.}

In tandem with radiometric data, magnetic data acquisition was carried out to assess the system's dual-sensing capabilities. The analysis employed the use of the 4th difference test. \textcolor{black}{The 4th difference is more appropriate and provides the finest resolution in comparison to the 2nd or 3rd difference tests which would be better suited for larger anomalies.} This analysis was aimed at identifying short-wavelength anomalies or high-frequency noise, thus ensuring the highest resolution of data. This method, by focusing on successive data differences, effectively highlighted values that deviated beyond a predefined threshold. Data collected at a base station over a brief period exhibited minor fluctuations, which were meticulously corrected during the data processing phase to account for diurnal variations \textcolor{black}{(the natural fluctuations in the Earth's magnetic field that occur over the course of a day due to solar and geomagnetic activity).} Comparing sensor specifications, the UAV's magnetometer boasted a wider measurement range and significantly higher sampling rate than the magnetometer used in the 2014 manned survey, highlighting the advancements in UAV-based mineral sensing technology.

To objectively assess the resolution difference between UAV and fixed-wing magnetometer data, the analysis focused on the pixel intensity distribution within the Total Magnetic Intensity (TMI) images, converted to grayscale for uniformity. Histograms, plotted on a logarithmic scale, were used to examine the grayscale value distribution of each image, leading to a quantifiable comparison through the calculation of the standard deviation of pixel intensities. The standard deviation for the fixed-wing TMI image was approximately 58.12 grayscale units, showcasing the range of pixel intensity variation. Conversely, the UAV TMI image demonstrated a slightly broader variation, with a standard deviation of around 60.82 grayscale units, indicating a wider array of intensity differences by approximately 2.7 grayscale units compared to its fixed-wing counterpart. This analysis underscores the UAV TMI's ability to capture a more diverse intensity range, suggesting a higher resolution in the magnetic data it collects.

Further comparative analysis between the UAV and the 2014 Navajo survey underscored the UAV's enhanced capability in identifying detailed magnetic anomalies, including their high and low variations, orientations, and spatial contacts seen in Fig.~\ref{fig:Gasdronemag.png}. Radiometric data from the UAV further delineated local variations in potassium and uranium, contrasting with broader regional trends observed in the fixed-wing survey data. In the combined magnetic and radiometric surveys, during the qualitative assessment of the two plots, the drone data depicted a similar trend within a ±10\% distribution of high and low potassium anomalies, with only slight variations in dynamic range values as seen in Fig.~\ref{fig:Rad vs MAG.png}. The integrated magnetic and radiometric survey analysis involved converting comparative graphs to grayscale to evaluate pixel intensity distributions independently. The fixed-wing survey graph displayed a standard deviation of about 61.24 grayscale units, reflecting its variability, while the UAV graph's standard deviation was slightly higher at approximately 67.84 grayscale units. This higher variability in the UAV data, by a margin of around 6.61 grayscale units, suggests a greater dynamic range in the drone-conducted surveys.

Statistical analysis of the multi sensor data acquisition datasets showed negligible differences in statistical parameters, indicating that conducting simultaneous magnetic and radiometric surveys does not detract from the quality of the radiometric data. This ability to concurrently gather magnetic and radiometric data without data interference underscores the effectiveness of integrated surveys in mineral exploration. The combined approach not only facilitates the understanding of surface compositions and alterations through radiometric data but also offers insights into deeper geological structures via magnetic data. 

\section{Conclusion}\label{sec:con}
This research has significantly progressed the field of mineral exploration by enhancing the capabilities of multirotor UAVs through the integration of advanced, high-sensitivity sensors. By addressing and mitigating issues such as electromagnetic interference and vibration noise, this study has markedly improved the accuracy of data collected during geological surveys. 

A notable breakthrough of this work is the development of a novel method for magnetometry, utilizing a suspension system that notably reduces noise levels compared to conventional fixed-mount configurations. Furthermore, the research has successfully integrated a sophisticated multi-sensor system on gas hybrid UAVs, which includes both radiometric and magnetometry sensors alongside a combined magnetometry and VLF sensor system. A pioneering 4-point mounting system was also introduced, significantly enhancing the stability of sensors, thereby elevating the quality of mineral exploration data acquired by UAVs. 
The findings of this study extend the utility of UAVs in mineral exploration, moving beyond mere aerial imagery to include high-precision, multi-faceted data collection. The demonstrated efficacy of these innovative solutions suggests a potential paradigm shift in methodologies employed for mineral data acquisition. The advancements presented herein not only underline the technical progress made but also underscore the increasing relevance and capability of UAVs in revolutionizing approaches to mineral exploration.





\bibliographystyle{IEEEtran}
\bibliography{./References.bib}

\begin{thebibliography}{10}
\providecommand{\url}[1]{#1}
\csname url@samestyle\endcsname
\providecommand{\newblock}{\relax}
\providecommand{\bibinfo}[2]{#2}
\providecommand{\BIBentrySTDinterwordspacing}{\spaceskip=0pt\relax}
\providecommand{\BIBentryALTinterwordstretchfactor}{4}
\providecommand{\BIBentryALTinterwordspacing}{\spaceskip=\fontdimen2\font plus
\BIBentryALTinterwordstretchfactor\fontdimen3\font minus \fontdimen4\font\relax}
\providecommand{\BIBforeignlanguage}[2]{{%
\expandafter\ifx\csname l@#1\endcsname\relax
\typeout{** WARNING: IEEEtran.bst: No hyphenation pattern has been}%
\typeout{** loaded for the language `#1'. Using the pattern for}%
\typeout{** the default language instead.}%
\else
\language=\csname l@#1\endcsname
\fi
#2}}
\providecommand{\BIBdecl}{\relax}
\BIBdecl

\bibitem{leech2021acquisition}
C.~Leech, S.~Burns, and K.~Hurley, ``{Acquisition Challenges for High Quality Data Using a UAV Deployed Magnetometer},'' in \emph{Sixth International Conference on Engineering Geophysics}, Virtual, 2021, oct. 25--28, 2021.

\bibitem{mohsan2023editorial}
S.~A.~H. Mohsan, M.~A. Khan, and Y.~Y. Ghadi, ``{Editorial on the Advances, Innovations and Applications of UAV Technology for Remote Sensing},'' \emph{Remote Sensing}, vol.~15, p. 5087, 2023.

\bibitem{vangu2022use}
G.~M. Vangu, ``The use of drones in mining operations,'' \emph{Mining Revue}, vol.~28, no.~3, pp. 73--82, 2022.

\bibitem{vitale2019new}
G.~Vitale, S.~Scudero, A.~D'Alessandro, A.~Pisciotta, R.~Martorana, and P.~Capizzi, ``New ultraportable data logger to perform magnetic surveys,'' in \emph{2019 International Symposium on Advanced Electrical and Communication Technologies (ISAECT)}, 2019.

\bibitem{persova2021resolution}
M.~G. Persova, Y.~G. Soloveichik, D.~V. Vagin, D.~S. Kiselev, A.~P. Sivenkova, and E.~I. Simon, ``{Resolution Analysis of Airborne Electromagnetic Survey Using Helicopter Platform and UAV},'' in \emph{2021 XV International Scientific-Technical Conference on Actual Problems Of Electronic Instrument Engineering (APEIE)}, Novosibirsk, Russian Federation, 2021, pp. 591--594.

\bibitem{jiang2020integration}
D.~Jiang, Z.~Zeng, S.~Zhou, Y.~Guan, and T.~Lin, ``{Integration of an Aeromagnetic Measurement System Based on an Unmanned Aerial Vehicle Platform and Its Application in the Exploration of the Ma’anshan Magnetite Deposit},'' \emph{IEEE Access}, vol.~8, pp. 189\,576--189\,586, 2020.

\bibitem{greengard2019when}
S.~Greengard, ``When drones fly,'' \emph{Communications of the ACM}, vol.~62, no.~11, pp. 16--18, 2019.

\bibitem{yang2022hybrid}
X.~Yang and X.~Pei, ``Hybrid system for powering unmanned aerial vehicles: Demonstration and study cases,'' in \emph{Hybrid Energy Systems: Hybrid Technologies for Power Generation}, M.~L. Faro, O.~Barbera, and G.~Giacoppo, Eds.\hskip 1em plus 0.5em minus 0.4em\relax Academic Press, 2022, pp. 439--473.

\bibitem{saeed2018survey}
A.~Saeed, A.~B. Younes, C.~Cai, and G.~Cai, ``A survey of hybrid unmanned aerial vehicles,'' \emph{Progress in Aerospace Sciences}, vol.~98, pp. 91--105, 2018.

\bibitem{barton2021extending}
I.~F. Barton, M.~J. Gabriel, J.~Lyons-Baral \emph{et~al.}, ``Extending geometallurgy to the mine scale with hyperspectral imaging: a pilot study using drone- and ground-based scanning,'' \emph{Mining, Metallurgy \& Exploration}, vol.~38, pp. 799--818, 2021.

\bibitem{niethammer2012uav}
U.~Niethammer, M.~R. James, S.~Rothmund, J.~Travelletti, and M.~Joswig, ``{{UAV}-based remote sensing of the Super-Sauze landslide: Evaluation and results},'' \emph{Engineering Geology}, vol. 128, pp. 2--11, 2012.

\bibitem{salvini2015geological}
R.~Salvini, S.~Riccucci, D.~Gullì, R.~Giovannini, C.~Vanneschi, and M.~Francioni, ``{Geological Application of {UAV} Photogrammetry and Terrestrial Laser Scanning in Marble Quarrying (Apuan Alps, Italy)},'' in \emph{Engineering Geology for Society and Territory - Volume 5}, G.~Lollino, A.~Manconi, F.~Guzzetti, M.~Culshaw, P.~Bobrowsky, and F.~Luino, Eds.\hskip 1em plus 0.5em minus 0.4em\relax Springer, 2015, pp. 1883--1887.

\bibitem{jackisch2019drone}
R.~Jackisch, Y.~Madriz, R.~Zimmermann, M.~Pirttijärvi, A.~Saartenoja, B.~H. Heincke, H.~Salmirinne, J.-P. Kujasalo, L.~Andreani, and R.~Gloaguen, ``{Drone-Borne Hyperspectral and Magnetic Data Integration: Otanmäki Fe-Ti-V Deposit in Finland},'' \emph{Remote Sensing}, vol.~11, no.~18, p. 2084, 2019.

\bibitem{shendryk2020fine}
Y.~Shendryk, J.~Sofonia, R.~Garrard, Y.~Rist, D.~Skocaj, and P.~Thorburn, ``{Fine-scale prediction of biomass and leaf nitrogen content in sugarcane using {UAV} LiDAR and multispectral imaging},'' \emph{International Journal of Applied Earth Observation and Geoinformation}, vol.~92, p. 102177, 2020.

\bibitem{goetz2009three}
A.~F. Goetz, ``{Three decades of hyperspectral remote sensing of the Earth: A personal view},'' \emph{Remote Sensing of Environment}, vol. 113, pp. S5--S16, 2009.

\bibitem{vosselman2013recognising}
G.~Vosselman, B.~Gorte, G.~Sithole, and T.~Rabbani, ``Recognising structure in laser scanner point clouds,'' in \emph{International Archives of the Photogrammetry, Remote Sensing and Spatial Information Sciences}, vol.~46, 2013, pp. 33--38.

\bibitem{thiele2021multi}
S.~T. Thiele, S.~Lorenz, M.~Kirsch, I.~C.~C. Acosta, L.~Tusa, E.~Herrmann, R.~Möckel, and R.~Gloaguen, ``{Multi-scale, multi-sensor data integration for automated 3-D geological mapping},'' \emph{Ore Geology Reviews}, vol. 136, p. 104252, 2021.

\bibitem{molnar2016unmanned}
A.~Molnar and C.~Parsons, ``{Unmanned Aerial Vehicles ({UAV}s) and Law Enforcement in Australia and Canada: Governance Through ‘Privacy’ in an Era of Counter-Law?}'' in \emph{National Security, Surveillance and Terror}, R.~Lippert, K.~Walby, I.~Warren, and D.~Palmer, Eds.\hskip 1em plus 0.5em minus 0.4em\relax Palgrave Macmillan, 2016, pp. 183--200.

\bibitem{eross2013three}
R.~Eröss, J.~B. Stoll, R.~Bergers, and B.~Tezkan, ``{Three-component VLF using an unmanned aerial system as sensor platform},'' \emph{First Break}, vol.~31, no.~7, 2013.

\bibitem{parshin2021lightweight}
A.~Parshin, Y.~Davidenko, S.~Yakovlev, V.~Vinokurov, and A.~Bashkeev, ``{Lightweight TEM and VLF Systems for Low-Altitude UAV-Based Geophysical},'' in \emph{European Meeting of Environmental and Engineering Geophysics}, 2021.

\bibitem{Halliday2021}
D.~Halliday, R.~Resnick, and J.~Walker, \emph{Fundamentals of Physics}, 12th~ed.\hskip 1em plus 0.5em minus 0.4em\relax John Wiley \& Sons, Inc., 2021.

\bibitem{walter2021optimization}
C.~A. Walter, ``{Optimization of UAV-Borne Aeromagnetic Surveying in Mineral Exploration},'' Ph.D. dissertation, Queen’s University, 2021, {Ph.D.} thesis.

\bibitem{usgs2023bipartisan}
{U.S. Geological Survey}, ``{Earth MRI Airborne Geophysical Survey Eon Geoscience Airplane with labels},'' \url{https://www.usgs.gov/media/images/plane-labelspng}, 2024, accessed: 2024-02-18.

\end{thebibliography}
%

\end{document}